\pdfoutput=1
\documentclass[11pt]{article}

\usepackage[]{acl}

\usepackage{times}
\usepackage{latexsym}

\usepackage[T1]{fontenc}
\usepackage[utf8]{inputenc}

\usepackage{microtype}

\usepackage{inconsolata}

\usepackage{pgfplots}
\usepgfplotslibrary{statistics}
\pgfplotsset{compat=1.18}

\usepackage{subcaption}
\usepackage{booktabs}
\usepackage{multirow}
\usepackage{array}
\usepackage{cellspace}
\usepackage{mathtools}

\title{
Measuring Cross-lingual Transfer in Bytes
}

\author{Leandro Rodrigues de Souza \\
  FEEC, UNICAMP, Brazil \\\And
  Thales Sales Almeida \\
  IC, UNICAMP, Brazil \\
  Maritaca AI, Brazil \\\AND
  Roberto Lotufo \\
  FEEC, UNICAMP, Brazil \\
  NeuralMind, Brazil\\\And
  Rodrigo Nogueira \\
  FEEC, UNICAMP, Brazil \\
  Maritaca AI, Brazil\\
  }

\begin{document}
\maketitle
\begin{abstract}
Multilingual pretraining has been a successful solution to the challenges posed by the lack of resources for languages. These models can transfer knowledge to target languages with minimal or no examples. Recent research suggests that monolingual models also have a similar capability, but the mechanisms behind this transfer remain unclear. Some studies have explored factors like language contamination and syntactic similarity. An emerging line of research suggests that the representations learned by language models contain two components: a language-specific and a language-agnostic component. The latter is responsible for transferring a more universal knowledge. However, there is a lack of comprehensive exploration of these properties across diverse target languages. To investigate this hypothesis, we conducted an experiment inspired by the work on the Scaling Laws for Transfer. We measured the amount of data transferred from a source language to a target language and found that models initialized from diverse languages perform similarly to a target language in a cross-lingual setting. This was surprising because the amount of data transferred to 10 diverse target languages, such as Spanish, Korean, and Finnish, was quite similar. We also found evidence that this transfer is not related to language contamination or language proximity, which strengthens the hypothesis that the model also relies on language-agnostic knowledge. Our experiments have opened up new possibilities for measuring how much data represents the language-agnostic representations learned during pretraining.\footnote{The code used in our experiments is publicly available at \url{https://github.com/lersouza/language-transfer}. We rely on the mC4 dataset from Huggingface, available at \url{https://huggingface.co/datasets/mc4}}
\end{abstract}

\section{Introduction}

The emergence of self-supervised pretraining models such as BERT has revealed a notable phenomenon of cross-lingual transfer even when these models are trained on multilingual corpora devoid of paired translation examples. For example, LLAMA~\cite{touvron2023llama}, which was trained self-supervisedly on an English-centric corpus, exhibits surprising multilingual capabilities~\cite{yuan2023multilingual,ye2023language}.
The underlying mechanisms driving this behavior remain unclear, with hypotheses ranging from the presence of shared ``anchor'' tokens~\cite{pires-etal-2019-multilingual} to language contamination~\cite{blevins-zettlemoyer-2022-language}, yet no scientific consensus has been reached.

Research in this area often involves the use of pre-existing language models (LMs), which are subsequently finetuned on supervised datasets in different languages \cite{desouza2021ability, yuan2023multilingual}. However, when evaluating multiple languages, conventional methodologies encounter two significant challenges: firstly, the dependence on supervised finetuning datasets, which often vary in size and quality, complicating cross-lingual comparisons; secondly, the use of subword tokenizers, which do not represent all languages equally.

In this work, we avoid these problems by working with a byte-level tokenizer and by using auto-regressive language models trained in self-supervised from scratch in one language and then finetuned on another.
To measure the effect of transfer learning, we employ the concept of data transfer~\cite{hernandez2021scaling}, which allows us to quantify how much each different source language contributes to the perplexity of the target language.

Our main contribution is providing a method that measures how much knowledge, in bytes, is transferred from one language to another. By applying it, our findings reveal a surprising trend: even when comparing linguistically distant languages, the data transfer metrics are of a comparable magnitude.
This research contributes additional evidence supporting the language-agnostic hypothesis, which suggests that the internal representations developed by a model are not only influenced by the linguistic surface form but also by the cultural and semantic content of the training data.

\section{Related Work}

Prior work attributed the success of multilingual models in cross-lingual transfer to ``anchor'' tokens~\cite{pires-etal-2019-multilingual}. However, subsequent research demonstrated that models could perform well even without these tokens~\cite{artetxe-etal-2020-cross}, highlighting the significance of shared parameters during training~\cite{conneau-etal-2020-emerging}. Competitive results were achieved by monolingual models with minimal or no adaptation~\cite{artetxe-etal-2020-cross,desouza2021ability}.

Investigations by \citet{blevins-zettlemoyer-2022-language} linked these findings to language contamination, where pretraining datasets contained target language data. Additional factors contributing to cross-lingual transfer success include dataset statistics, language attributes~\cite{lin-etal-2019-choosing}, language structure~\cite{lin-etal-2019-choosing,papadimitriou-jurafsky-2020-learning,pretrainwithouthuman,ri-tsuruoka-2022-pretraining}, and token overlap between training and target languages~\cite{beukman2023analysing}. The role of language script~\cite{fujinuma-etal-2022-match} and model tokenizer~\cite{rust-etal-2021-good} was also noted, prompting the use of a byte tokenizer to address these issues~\cite{xue-etal-2022-byt5,abonizio-etal-2022-monobyte}.

Recent research proposed a two-component model representation hypothesis—language agnostic and language specific~\cite{desouza2021ability,zeng-etal-2023-soft,wu-etal-2022-laft}. While promising, no study has measured how much of the language-agnostic component is used in settings with multiple source and target languages. Additionally, existing research still applies the source language vocabulary to the target language, potentially compromising input representations and affecting results.

To address these gaps, we draw on \citet{hernandez2021scaling} and employ a byte vocabulary in our experiments to overcome current literature limitations.

\section{Methodology}

Inspired by~\citet{hernandez2021scaling}, our methodology focuses on quantifying the transferability of pretraining data across distributions, particularly between different languages. We select a \textbf{target} language and finetune models initialized from various \textbf{source} languages onto it. Subsequently, we evaluate each model on a \textbf{target} language test set and compare their performance. We introduce the Data Transfer ($D_T$) metric to estimate knowledge transfer, explained in more depth in \ref{sec:methodology-dt}.
This process is repeated across different \textbf{target} languages to observe effects across a broad linguistic spectrum.

These experiments aim to quantify and compare cross-lingual knowledge transfer from different source languages. This analysis seeks to uncover the extent to which transferability depends on specific languages and the importance of language-agnostic components in learned representations. The following sections provide further details.

\subsection{Data Transfer Estimation} \label{sec:methodology-dt}

\begin{figure*}[htbp]
    \centering
    \begin{tikzpicture}
            \begin{axis}[
                title style={at={(0.5,1.1)}, anchor=center},
                title=Target Language: Spanish,
                xlabel={Finetuning Dataset (Spanish) Size in Bytes},
                ylabel={Test Set PPL in Spanish},
                legend pos=north east,
                grid=none,
                width=0.7\textwidth,
                height=6cm,
                xmode=log,
                log basis x=e,
            ]
            
            % Plot 1: Scratch
            \addplot[
                color=black,
                mark=*,
            ]
            coordinates {
                (6000000, 12.539025)
                (19000000, 9.372363)
                (60000000, 3.570520)
                (189000000, 2.850784)
                (600000000, 2.546002)
                (6000000000, 2.668534)
            };
            \addlegendentry{From scratch}
    
            % Plot 2: Initialization 2
            \addplot[
                color=blue,
                mark=square*,
            ]
            coordinates {
                (6000000, 3.163111)
                (19000000, 2.943550)
                (60000000, 2.741453)
                (189000000, 2.581411)
                (600000000, 2.408924)
                (6000000000, 2.390901)
            };
            \addlegendentry{From English}

            \end{axis}

            \draw[<-|, gray] (0.2,1.8)--(4.2,1.8) node[midway, above] {$D_E$};
            \draw[<-|, gray] (0.2,0.65)--(0.7,0.65) node[midway, above] {$D_F$};
            \draw[|-|, gray] (0.9,0.65)--(4.2,0.65) node[midway, above] {$D_T$};
        \end{tikzpicture}
    \caption{Example illustrating how the coeficients $D_T$, $D_F$ and $D_E$ are calculated. Each series represents a different initialization. $D_T$ is the number of additional tokens in the target language that a from-scratch model would have needed to achieve the same perplexity of a model finetuned from English. $D_F$ is the size of the dataset used for finetuning and $D_E$ accounts for all data, both $D_F$ and $D_T$.}
    \label{fig:explain-dt}
\end{figure*}
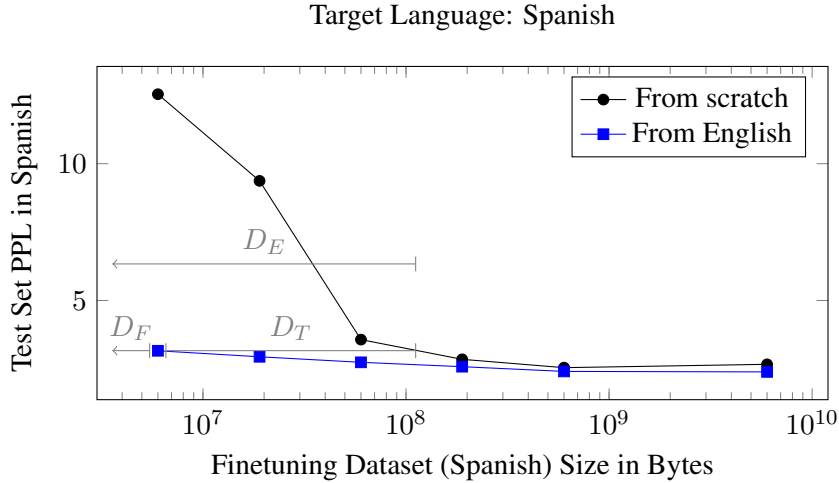

To quantify the knowledge transfer from pretrained models to a given target language, we utilize the Data Transfer ($D_T$) metric. This metric assesses the effectiveness of pretraining data by measuring the additional tokens required in the target language for a model initialized from scratch to match the performance of a model pretrained on a source language and finetuned on the target. Figure~\ref{fig:explain-dt} illustrates this concept.

%In our experiments, we utilize a set of $M$ different dataset sizes, denoted as $\mathbf{D_F} = \{s_0, s_1, ..., s_m\}$, in the target language. Initially, we train a random-initialized model on these datasets, resulting in a set of perplexities denoted as $\mathbf{P_R}$, encompassing values $p_{r,0}, p_{r,1}, ..., p_{r,m}$. Subsequently, we finetune on the target language a pretrained model in a source language $\ell$ using the same dataset sizes $\mathbf{D_F}$, generating another set of perplexities denoted as $\mathbf{P_\ell}$, with perplexity values $p_{\ell,0}, p_{\ell,1}, ..., p_{\ell,m}$.

In our experiments, we utilize a set of $M$ different dataset sizes, denoted as $\mathbf{D_F} = \{s_0, s_1, ..., s_m\}$, in the target language. Initially, we train a random-initialized model on these datasets, resulting in a set of perplexities $\mathbf{P_R} = \{p_{r,0}, p_{r,1}, ..., p_{r,m}\}$. Subsequently, we train from scratch another model on a fixed amount of tokens (e.g., 6B) in a source language $\ell$, and then finetune it on the target language using the same dataset sizes $\mathbf{D_F}$, generating another set of perplexities $\mathbf{P_\ell} = \{p_{\ell,0}, p_{\ell,1}, ..., p_{\ell,m}\}$.

To estimate the Data Effective metric ($D_E$), which represents the amount of data needed to achieve a certain performance, we utilize a linear interpolation function $\gamma(y', \mathbf{X}, \mathbf{Y})$. This function interpolates between discrete data points $(x_j, y_j) \in X \times Y$, evaluated at $y'$. In our case, the calculation is expressed as:

\begin{equation}
    D_{E,i} = \gamma(p_{\ell,i},\, \mathbf{D_F},\, \mathbf{P_R}) 
\end{equation}

Here, $D_{E,i}$ signifies the Data Effective metric for the $i$-th perplexity value in $\mathbf{P_\ell}$. $\mathbf{P_R}$ denotes the set of perplexity values derived from the random-initialized model, while $\mathbf{D_F}$ represents the dataset sizes employed during finetuning.

We utilize the linear interpolation function provided by the NumPy library to approximate $D_{E,i}$. Further details can be found in the NumPy documentation.\footnote{See \url{https://numpy.org/doc/stable/reference/generated/numpy.interp.html}}

Finally, the Data Transfer metric is computed by subtracting the $i$-th dataset size $s_i$ from $D_{E,i}$:

\begin{equation}
    D_{T,i} = D_{E,i} - s_i
    \label{eq:data_transfer_formula}
\end{equation}

By subtracting the dataset size from $D_{E,i}$, we account only for the data coming from pretraining. Since a byte vocabulary is utilized, the amount of data transferred is measured in \textbf{bytes}.

\subsection{Task and Evaluation Metric}

We adopt Language Modeling as our main task with perplexity as the performance metric for all experiments. Perplexity, derived from the model's loss ($e^{loss}$), facilitates future predictions of model behavior in transfer learning scenarios, following the approach in \citet{hernandez2021scaling}. This choice also allows extensive experiments across multiple languages, leveraging datasets like mC4~\cite{xue-etal-2021-mt5} to overcome size limitations inherent in supervised datasets.

\subsection{Tokenization Impact}

In cross-lingual setups, the choice of tokenization method holds considerable significance \cite{rust-etal-2021-good}. While subword tokenizers are commonly employed in cross-lingual experiments, using a tokenizer trained in a source language on a distant target language may result in an increased number of tokens. This can lead to the utilization of undertrained embeddings in some instances, introducing challenges for effective sentence representation. Furthermore, dealing with different scripts introduces the issue of numerous ``unknown'' tokens, exacerbating the difficulty of obtaining suitable input representations for the model.

To address these challenges, we opt for a byte vocabulary based on the approach proposed by \citet{xue-etal-2022-byt5}, which allows us to standardize representations across all languages, ensuring that each model encounters the same quantity of UTF-8 bytes. By doing so, we mitigate the use of unknown tokens and undertrained embeddings, thereby minimizing the impact of tokenization issues on the performance of our experiments.

\subsection{Language Contamination} 
A potential reason for a pretrained model's superior performance in cross-lingual tasks is the presence of a substantial amount of data in the target language in its pretraining dataset, a phenomenon referred to as language contamination. To quantify this impact, following the approach outlined by \citet{blevins-zettlemoyer-2022-language}, we examine the rates of target language fragments in the source language dataset and vice versa.

Specifically, for a given source language $\ell$ and target language $t$, we calculate the ratio of all lines classified as $\ell$ in the target dataset (known as contamination on target) and as $t$ in the pretraining dataset (known as contamination on source). We perform language detection using the \textit{fasttext} tool \citep{bojanowski2017enriching}, employing a threshold of 0.6 for classification.

Next, we compute the Spearman correlation between the set of Data Transfer metrics $\mathbf{D_T}$ and those ratios obtained from the outcomes of our experiments.

Correlating these rates with the model's data transfer indicator allows us to evaluate the impact of language contamination on model performance.

\subsection{Language Similarity}

Language similarity is often cited as a crucial factor influencing cross-lingual transfer performance in natural language processing tasks. In this work, we aim to investigate the relationship between language similarity and cross-lingual transfer effectiveness based on the outcomes of our experiments.

To explore this relationship, we measure various distances between languages, including syntactic, geographic, and phonological distances. These distances are calculated based on the methodology proposed by \citet{littell-etal-2017-uriel}.

We aim to correlate these language distances with the data transfer metric ($D_T$). By employing Spearman correlation analysis, we seek to discern whether there exists a significant correlation between the observed $D_T$ values and the measured language distances.

This analysis elucidates whether our experimental results can be attributed to the similarities between the languages involved in our cross-lingual experiments.

\section{Experiments}

This section presents the languages, datasets, model architecture, and training details for our experiments.

\subsection{Languages}

\medskip
\noindent\textbf{Source Languages Selection}. We chose three diverse languages—English, Russian, and Chinese—for the source language during the pretraining phase. This selection ensures a broad linguistic spectrum while adhering to pretraining budget constraints.

\medskip
\noindent\textbf{Target Languages Selection}. Ten target languages, spanning various language families and different scripts, were chosen to establish a diverse cross-lingual setting. Details, including language codes, are provided in Table \ref{tab:languages}.

\begin{table}
    \centering\resizebox{0.45\textwidth}{!}{
    \begin{tabular}{cccc}
        \toprule
        \textbf{Code} & \textbf{Language} & \textbf{Family} & \textbf{Script} \\
        \midrule
        ar & Arabic & Afro-Asiatic & Arabic \\
        en & English & Indo-European & Latin \\
        es & Spanish & Indo-European & Latin \\
        zh & Chinese & Sino-Tibetan & Hanzi \\
        fi & Finnish & Uralic & Latin \\
        de & German & Indo-European & Latin \\
        ko & Korean & Koreanic & Hangul \\
        id & Indonesian & Austronesian & Latin \\
        ja & Japanese & Japonic & Kanji, Hiragana, Katakana \\
        ru & Russian & Indo-European & Cyrillic \\
        \bottomrule
    \end{tabular}
    }
    \caption{Characteristics of selected target languages.}
\label{tab:languages}
\end{table}

\subsection{Datasets}

For training and finetuning, language subsets from the mC4 dataset~\cite{xue-etal-2021-mt5} for the selected languages were utilized.\footnote{See \url{https://huggingface.co/datasets/mc4} for more details.} Pretraining datasets comprised approximately 6 billion tokens, while finetuning datasets ranged from 6 million to 6 billion tokens. Documents were sampled at random without replacement until the desired amount of tokens was reached.

\subsection{Model Architecture}

Our model is a decoder-only Transformer~\cite{vaswani2017attention} that uses a byte vocabulary with 256 embeddings with a dimension of 640. The model follows closely the implementation provided in the T5x library. It consists of 10 layers, each having 10 attention heads with dimensions of 64. The intermediate dimension of Multi-Layer Perceptron (MLP) has a dimension of 2560 and GELU \cite{hendrycks2023gaussian} activations. The parameters of the embeddings matrix and the final dense layer are shared. We use relative positional embeddings~\cite{shaw2018self}. The resulting model has approximately 65 million parameters.

\subsection{Training details}

Models were trained using a causal language modeling objective. Each batch has 512 sequences of 1024 tokens. We use the AdamW optimizer with an initial learning rate of 2e-4, which decayed to 2e-5 through cosine decay following \citet{hoffmann2022training}. Finetuning employed a constant learning rate of 2e-5 over 10 epochs, except for the 6 billion dataset size where we limited it to 3 epochs. This adjustment was based on preliminary experiments indicating that the model tends to overfit beyond this epoch count in larger datasets. The best model was selected based on the lowest perplexity achieved on the development set. Warmup steps varied with finetuning dataset sizes (ranging from 0 for smaller datasets to 3000 for larger ones), aligning with findings that smaller datasets completed finetuning before warmup completion \cite{hernandez2021scaling}. We utilized the T5X framework \cite{roberts2022t5x} for our experiments. We used a total of 600 hours of a TPU v2-8 (seven hours of pretraining per model, and fifteen hours for the largest finetuning).

\section{Results}

Results are compiled in Table~\ref{tab:results-table}, where we exclusively report instances involving different source and target languages.

Given our methodology, where we vary the source language while keeping the target language constant to assess the impact of pretraining language in cross-lingual scenarios, it is essential to read the table vertically unless stated otherwise. Each row represents the results obtained by finetuning the model from a specific source language to a given target language (indicated in the column). This vertical arrangement facilitates the comparison of model performance across different source languages for the same target language.
Furthermore, test sets vary significantly for each language due to the nature of the mC4 dataset. Consequently, results across target languages are not comparable.

Throughout this section, we highlight findings from models finetuned on 6 million tokens (i.e., $\mathbf{D_F}=\{\textrm{6MB}\}$) unless otherwise specified.\footnote{This restriction applies only to the finetuned models. For the from-scratch ones, we need their perplexities on multiple target language dataset sizes to estimate $D_T$.} This extreme scenario tests models with minimal target language resources.

\begin{table*}[htbp]
    \small
    \centering
    \begin{tabular}{ccccccccccccc}
        \toprule
        \textbf{Source Lang.} & \textbf{Metric} & \textbf{ar} & \textbf{de} & \textbf{en} & \textbf{es} & \textbf{fi} & \textbf{id} & \textbf{ja} & \textbf{ko} & \textbf{ru} & \textbf{zh} \\
        \midrule
        Scratch init. & Perplexity & 6.44 & 14.82 & 16.28 & 12.54 & 12.71 & 12.00 & 12.47 & 11.69 & 6.27 & 15.34 \\
        \midrule
        \multirow{2}{*}{English} & Perplexity & 2.82 & 3.67 & - & 3.16 & 3.57 & 2.61 & 3.92 & 3.58 & 2.44 & 4.43 \\
        & $D_{T}$ & \textbf{101.02} & \textbf{95.25} & - & \textbf{121.14} & \textbf{76.57} & \textbf{102.62} & 47.50 & 48.74 & \textbf{75.64} & \textbf{29.21} \\
        \midrule
        \multirow{2}{*}{Russian} & Perplexity & 2.83 & 3.98 & 3.66 & 3.47 & 3.80 & 2.84 & 3.89 & 3.58 & - & 4.52 \\ 
        & $D_{T}$ & 99.00 & 47.87 & \textbf{174.63} & 67.88 & 50.96 & 51.32 & 47.81 & 48.69 & - & 26.18 \\
        \midrule
        \multirow{2}{*}{Chinese} & Perplexity & 2.88 & 4.26 & 3.89 & 3.75 & 3.98 & 2.98 & 3.46 & 3.48 & 2.72 & - \\
        & $D_{T}$ & 90.63 & 31.76 & 66.96 & 50.27 & 49.65 & 50.21 & \textbf{69.48} & \textbf{49.88} & 48.47 & - \\
        \bottomrule
    \end{tabular}
    \caption{Results for Perplexity and Data Transfer (in MB) for all target and source languages. All metrics are reported after finetuning the models in 6 million tokens of the target language.}
    \label{tab:results-table}    
\end{table*}

\subsection{Performance with different initializations}

\begin{figure*}
    \begin{subfigure}[b]{0.33\textwidth}
        \begin{tikzpicture}
            \begin{axis}[
                title style={at={(0.5,1.1)}, anchor=center},
                title=Target: Spanish,
                xlabel={Finetuning Dataset Size},
                ylabel={Perplexity},
                legend pos=north east,
                grid=major,
                width=1.0\textwidth,
                height=5cm,
                xmode=log,
                log basis x=e,
            ]
            
            % Plot 1: Scratch
            \addplot[
                color=black,
                mark=*,
            ]
            coordinates {
                (6000000, 12.539025)
                (19000000, 9.372363)
                (60000000, 3.570520)
                (189000000, 2.850784)
                (600000000, 2.546002)
                (6000000000, 2.668534)
            };
            \addlegendentry{scratch}
    
            % Plot 2: Initialization 2
            \addplot[
                color=blue,
                mark=square*,
            ]
            coordinates {
                (6000000, 3.163111)
                (19000000, 2.943550)
                (60000000, 2.741453)
                (189000000, 2.581411)
                (600000000, 2.408924)
                (6000000000, 2.390901)
            };
            \addlegendentry{en}

            % Plot 2: Initialization 3
            \addplot[
                color=yellow,
                mark=square*,
            ]
            coordinates {
                (6000000, 3.752540)
                (19000000, 3.519423)
                (60000000, 2.987010)
                (189000000, 2.692495)
                (600000000, 2.455862)
                (6000000000, 2.445876)
            };
            \addlegendentry{zh}

            % Plot 2: Initialization 4
            \addplot[
                color=red,
                mark=square*,
            ]
            coordinates {
                (6000000, 3.474664)
                (19000000, 3.133820)
                (60000000, 2.855067)
                (189000000, 2.634654)
                (600000000, 2.419389)
                (6000000000, 2.271673)
            };
            \addlegendentry{ru}
            \end{axis}
        \end{tikzpicture}
    \end{subfigure}\hspace{10mm}%
    \begin{subfigure}[b]{0.33\textwidth}
        \begin{tikzpicture}
            \begin{axis}[
                title style={at={(0.5,1.1)}, anchor=center},
                title=Arabic,
                xlabel={Finetuning Dataset Size},
                ylabel={},
                ymin=0,
                ymax=15,
                ytick={5,10},
                legend pos=north west,
                grid=major,
                width=1.0\textwidth,
                height=5cm,
                xmode=log,
                log basis x=e,
            ]
            
            % Plot 1: Scratch
            \addplot[
                color=black,
                mark=*,
            ]
            coordinates {
                (6000000, 6.443963)
                (19000000, 5.398603)
                (60000000, 3.111124)
                (189000000, 2.384848)
                (600000000, 2.105888)
                (6000000000, 1.929630)
            };
            \addlegendentry{scratch}
    
            % Plot 2: Initialization 2
            \addplot[
                color=blue,
                mark=square*,
            ]
            coordinates {
                (6000000, 2.818745)
                (19000000, 2.573541)
                (60000000, 2.382326)
                (189000000, 2.169670)
                (600000000, 2.025625)
                (6000000000, 2.141356)
            };
            \addlegendentry{en}

            % Plot 2: Initialization 3
            \addplot[
                color=yellow,
                mark=square*,
            ]
            coordinates {
                (6000000, 2.880094)
                (19000000, 2.678952)
                (60000000, 2.461752)
                (189000000, 2.208565)
                (600000000, 2.059278)
                (6000000000, 2.082330)
            };
            \addlegendentry{zh}

            % Plot 2: Initialization 4
            \addplot[
                color=red,
                mark=square*,
            ]
            coordinates {
                (6000000, 2.830716)
                (19000000, 2.614603)
                (60000000, 2.438054)
                (189000000, 2.172620)
                (600000000, 2.031723)
                (6000000000, 1.930613)
            };
            \addlegendentry{ru}
            \legend{};
            \end{axis}
        \end{tikzpicture}
    \end{subfigure}%
    \begin{subfigure}[b]{0.33\textwidth}
        \begin{tikzpicture}
            \begin{axis}[
                title style={at={(0.5,1.1)}, anchor=center},
                title=Japanese,
                xlabel={Finetuning Dataset Size},
                ylabel={},
                legend pos=north west,
                grid=major,
                width=1.0\textwidth,
                height=5cm,
                xmode=log,
                log basis x=e,
            ]
            
            % Plot 1: Scratch
            \addplot[
                color=black,
                mark=*,
            ]
            coordinates {
                (6000000, 12.468638)
                (19000000, 7.148220)
                (60000000, 3.551774)
                (189000000, 2.900728)
                (600000000, 2.540472)
                (6000000000, 2.277216)
            };
            \addlegendentry{scratch}
    
            % Plot 2: Initialization 2
            \addplot[
                color=blue,
                mark=square*,
            ]
            coordinates {
                (6000000, 3.919586)
                (19000000, 3.322470)
                (60000000, 2.993006)
                (189000000, 2.706201)
                (600000000, 2.465089)
                (6000000000, 2.544193)
            };
            \addlegendentry{en}

            % Plot 2: Initialization 3
            \addplot[
                color=yellow,
                mark=square*,
            ]
            coordinates {
                (6000000, 3.456637)
                (19000000, 3.155288)
                (60000000, 2.888835)
                (189000000, 2.649093)
                (600000000, 2.427642)
                (6000000000, 2.400873)
            };
            \addlegendentry{zh}

            % Plot 2: Initialization 4
            \addplot[
                color=red,
                mark=square*,
            ]
            coordinates {
                (6000000, 3.891109)
                (19000000, 3.382605)
                (60000000, 3.023181)
                (189000000, 2.719937)
                (600000000, 2.454615)
                (6000000000, 2.308033)
            };
            \addlegendentry{ru}
            \legend{};
            \end{axis}
        \end{tikzpicture}
    \end{subfigure}
    \caption{Results measured in Perplexity per token for three target languages. Each series represents a different initialization: train from scratch, finetune from an English, Chinese, or Russian model.}
    \label{fig:results-in-perplexity}
\end{figure*}
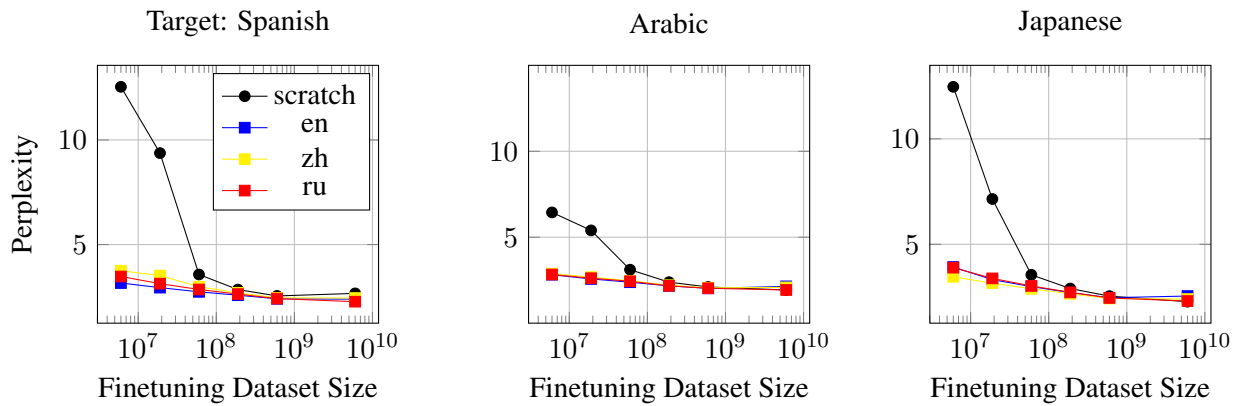

We delve into the results of three target languages: Spanish, Arabic, and Japanese. The perplexity scores across all dataset sizes for these languages are highlighted in Figure~\ref{fig:results-in-perplexity}, offering a chance for in-depth analysis despite the constraints of space.

A key observation is the consistent proximity of perplexity values for all three source languages in every target language. For instance, while one might expect a significant advantage for Chinese as a source language when finetuned in Japanese, or for English when paired with Spanish, this is not the case. This suggests the model leverages source language representations even when it lacks significant similarity with the target language.

Taken together, our results indicate that the model can rely on representations beyond those capturing language structure. This observation supports the recent hypothesis that these representations encompass both language-specific and language-neutral components, strengthening the latter as an important aspect.

\subsection{Data Transfer estimation for target languages}

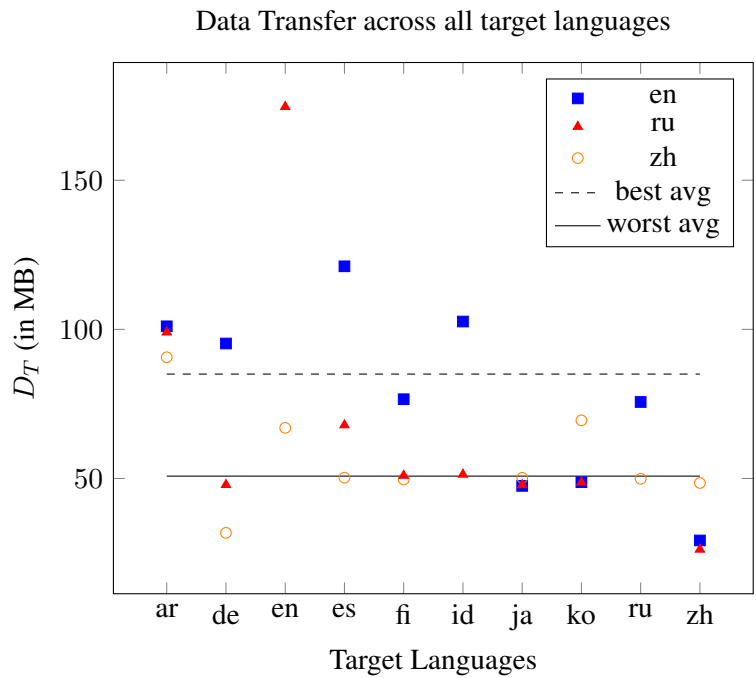
\begin{figure*}
    \centering
    \begin{tikzpicture}
        \begin{axis}[
            title={Data Transfer across all target languages},
            width=10cm,
            xlabel={Target Languages},
            ylabel={$D_T$ (in MB)},
            legend pos=north east,
            xtick={1, 2, 3, 4, 5, 6, 7, 8, 9, 10},
            xticklabels={ar, de, en, es, fi, id, ja, ko, ru, zh},
            scatter/classes={
                en={mark=square*,blue},
                ru={mark=triangle*,red},
                zh={mark=o,orange}
            },
        ]

         % English data
         \addplot[scatter,only marks,scatter src=explicit symbolic] coordinates {
             (1, 101.02) [en]
             (2,  95.25) [en]
             (4, 121.14) [en]
             (5,  76.57) [en]
             (6, 102.62) [en]
             (7,  47.50) [en]
             (8,  48.74) [en]
             (9,  75.64) [en]
             (10, 29.21) [en]
             (1, 90.63) [zh]
             (2, 31.76) [zh]
             (3, 66.96) [zh]
             (4, 50.27) [zh]
             (5, 49.65) [zh]
             (7, 50.21) [zh]
             (8, 69.48) [zh]
             (9, 49.88) [zh]
             (10, 48.47) [zh]
            (1,  99.00 )  [ru]
            (2,  47.87 )  [ru]
            (3,  174.63)  [ru]
            (4,  67.88 )  [ru]
            (5,  50.96 )  [ru]
            (6,   51.32)  [ru]
            (7,   47.81)  [ru]
            (8,   48.69)  [ru]
            (10,   26.18)  [ru]
        };
            \addplot[dashed, black] coordinates {(1, 85.00) (10, 85.00)};
        
            \addplot[solid, black] coordinates {(1, 50.78) (10, 50.78)};
        
            \legend{en, ru, zh, best avg, worst avg}
        \end{axis}
    \end{tikzpicture}
    \caption{Dispersion chart for Data Transfer ($D_T$) across target languages. Each series corresponds to a distinct source language. The first dashed line (top-to-bottom) indicates the average of the best results (higher transfer), while the second one represents the average of the worst results (lower transfer).}
    \label{fig:dispersion-chart}
\end{figure*}

\begin{figure}
    \begin{tikzpicture}
      \begin{axis}[
        boxplot/draw direction=y,
        % ymode=log,
        % log basis y={e},
        xtick={1,2,3},
        xticklabels={en, ru, zh},
        xlabel={Source Languages},
        ylabel={$D_T$ (in MB)},
        boxplot/box extend=0.5,
        width=7cm,
        height=4cm,
        ymin=0
      ]
    
      % Series for en
        \addplot+[boxplot] table[y index=0] {
            105.931 
            99.87413 
            127.0209 
            80.29443 
            107.6085 
            49.80689 
            51.10478 
            79.31563 
            30.62883 
        };
    
      % Series for ru
      \addplot+[boxplot] table[y index=0] {
            103.8057
            50.19782
            183.1094
            71.18048
            53.43318
            53.80960
            50.13154
            51.05506
            27.45323
      };
    
      % Series for zh
      \addplot+[boxplot] table[y index=0] {
            95.03516
            33.30020
            70.21128
            52.71371
            52.06323
            52.64918
            72.85078
            52.30758
            50.82234
      };
    
      \end{axis}
    \end{tikzpicture}
    \caption{Boxplot with Data Transfer results for the 6 million tokens datasets in all target languages.}
    \label{fig:boxplot-dt-source-languages}
\end{figure}
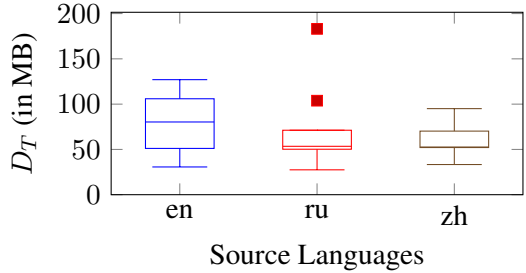

Analysis of the Data Transfer ($D_T$) metric in Table~\ref{tab:results-table} reveals that values are consistently close across different source languages for a given target. Notable examples include \textbf{Arabic} (en: 101MB, ru: 99MB, zh: 90MB), \textbf{Japanese} (en: 47.5MB, ru: 47.8MB, zh: 69.48MB), and \textbf{Finnish} (en: 102.62MB, ru: 51.32MB, zh: 76.57MB).

A clear pattern emerges: one initialization stands out for most target languages, while the other two show similar $D_T$ values. This pattern is visually represented in Figure~\ref{fig:dispersion-chart}, particularly evident for Finnish (fi), Indonesian (id), Japanese (ja), Korean (ko), and Chinese (zh). Additionally, closely clustered results are observed for Arabic (ar), German (de), Spanish (es), and Russian (ru).

With three diverse initializations, the fact that two consistently show similar $D_T$ values, even when distant from the target language (e.g., Russian and Chinese for Finnish), suggests the models leverage language-agnostic representations. This observation aligns with the expected behavior, where $D_T$ would be close to zero if language-specific knowledge were dominant.

English (en) demonstrates effective knowledge transfer to most target languages, potentially due to its widespread presence in corpora across languages. This hypothesis is explored further in Section~\ref{section:lang-contamination}. Additionally, Chinese (zh) appears to transfer effectively to Japanese (ja) and Korean (ko), likely due to their linguistic proximity.

We observe that Chinese (zh) tends to transfer effectively to Japanese (ja) and Korean (ko), both of which are considered closer languages. This, together with our other observations, indicates that, while not determinant, language-specific component also plays a role in cross-lingual transfer.

Finally, upon examination of Figure~\ref{fig:boxplot-dt-source-languages}, we also observe that most $D_T$ values are clustered between 50MB and 100MB, with low variation within a source language across all target languages. While not ideal, this comparison across target languages indicates that the transfer from our source models is relatively consistent, not spanning more than one order of magnitude.

\subsection{Language Contamination Impact} \label{section:lang-contamination}

Table~\ref{tab:correlation-contam} summarizes the results of assessing the language contamination effect in our experiments.

\begin{table}[htbp]
    \small
    \centering
    \begin{tabular}{|c|c|c|}
        \hline
        \textbf{Correlation} & $\rho$ & \textit{p-value} \\
        \hline
        $\mathbf{D_T}$ and contamination on source & 0.191 & 0.0157 \\
        $\mathbf{D_T}$ and contamination on target & 0.265 & 0.0021 \\
        \hline
    \end{tabular}
    \caption{Spearman Correlation ($\rho$) and p-value assessing the correlation of $\mathbf{D_T}$ with both the ratio of a target language in the source dataset (contamination on source) and with source language in the target dataset (contamination on target).}
    \label{tab:correlation-contam}
\end{table}

The analysis excludes the 6 billion tokens finetuning dataset size to mitigate the ossification effect, as observed in \citet{hernandez2021scaling}. This effect leads to a performance drop for pretrained models with larger finetuning datasets, worsening perplexity compared to scratch-trained models. Excluding this data point helps avoid introducing noise and adverse effects on coefficient calculation, given its singular occurrence per source-target language pair. Additionally, due to the sample size (< 500 observations), the permutation test is utilized to calculate the \textit{p-value}.

Although a correlation of 0.191 exists between $D_T$ and contamination on the source dataset, this coefficient indicates a weak association, suggesting a minimal impact on cross-lingual performance, contradicting the findings by \citet{blevins-zettlemoyer-2022-language}.

Exploration of language contamination in target datasets reveals a higher correlation of 0.265, particularly influenced by widespread languages like English. However, this coefficient still signifies a weak association between $D_T$ and target contamination, thus not supporting the language contamination hypothesis.

\subsection{Language Similarity and Data Transfer} \label{section:lang-distance}

This subsection outlines our analysis of the correlation between language distances and the data transfer metric ($D_T$), summarized in Table~\ref{tab:correlation-similarity}.

\begin{table}[htbp]
    \small
    \centering
    \begin{tabular}{|c|c|c|}
        \hline
        \textbf{Measure to correlate with $D_T$} & $\rho$ & \textit{p-value} \\
        \hline
        Syntactic distance & -0.147 & \multirow{6}{3em}{$> 0.7$} \\
        Geographic distance & -0.110 & \\
        Phonological distance & -0.117 & \\
        Genetic distance & -0.150 & \\
        Inventory distance & -0.090 & \\
        Featural distance & -0.145 & \\
        \hline
    \end{tabular}
    \caption{Spearman Correlation ($\rho$) and p-value assessing the correlation of $\mathbf{D_T}$ with a diverse set of language distance measurements.}
    \label{tab:correlation-similarity}
\end{table}

We find a weak correlation between source-target language distances and Data Transfer. Since we are considering multiple language characteristics, such as syntax and phonology, the results suggest that language similarity has a minor role in knowledge transfer between distinct languages. Nonetheless, the small number of source languages necessitates a cautious interpretation of these results, especially since all obtained p-values exceed 0.7, indicating limited statistical significance.

Because of that, we conducted a controlled experiment, pretraining a language model in Portuguese and evaluating its performance on the Spanish target language --- a language known for its similarity to Portuguese. Results, depicted in Table~\ref{tab:results-table-portuguese}, were compared across various initializations, including more distant languages like Chinese.

\begin{table}[htbp]
    \small
    \centering
    \begin{tabular}{|c|cc|}
        \hline
        & \multicolumn{2}{|c|}{\textbf{Spanish}}\\
        \textbf{Source Lang.} & \textbf{$D_T$} & Perplexity
        \\
        \hline
        \textbf{Portuguese} & \textbf{164.47} & \textbf{2.91} \\
        English & 121.14 & 3.16\\
        Russian & 67.88 & 3.47 \\
        Chinese & 50.27 & 3.75 \\
        \hline
    \end{tabular}
    \caption{Results for Data Transfer (in MB) and Perplexity in Spanish, highlighting Portuguese as a source language. All metrics are reported after finetuning the models in 6 million tokens of the target language. English, Russian, and Chinese results are the same as Table~\ref{tab:results-table}, added to facilitate comparison.}
    \label{tab:results-table-portuguese}    
\end{table}

When initialized with Portuguese, the model achieves a lower Perplexity in Spanish compared to when initialized with other languages. Additionally, $D_T$ peaks among all initializations, suggesting the influence of language proximity between Portuguese and Spanish. English initialization also yields comparable results, with a $D_T$ difference of around 40 MB and a perplexity variation of 0.25. Chinese and Russian show the lowest, yet similar, scores.

One possible interpretation is that the language-agnostic component accounts for 50\% of the transfer, with Russian and Chinese being more distant from Spanish, while the language-specific component contributes the remaining 50\%, considering closer linguistic and script systems. However, further investigation with more language pairs is necessary to determine the actual factors influencing transfer performance, including linguistic structure, script, or shared cultural knowledge in pretraining datasets.

\subsection{Commutative property exploration}
\begin{table}[htbp]
    \small
    \centering
    \begin{tabular}{|c|c|c|c|}
        \hline
         \textbf{Pair ($L_1$, $L_2$)} & $L_1 \rightarrow  L_2$ & $L_2 \rightarrow  L_1$ & $\Delta$  \\
         \hline
        en, ru & 75.64 & 174.63 & 98.99\\
        en, zh & 29.21 & 66.96 & 37.75\\
        ru, zh & 26.18 & 48.47 & 22.29\\
         \hline
    \end{tabular}
    \caption{Analysis of the Commutative Property in terms of Data Transfer $D_T$. We analyze pairs of languages ($L_1$, $L_2$), reporting the observed $D_T$ from $L_1$ to $L_2$ and vice-versa. Values are reported in megabytes.}
    \label{tab:commutative-property}
\end{table}

We examine the commutative property of data transfer between English (en), Russian (ru), and Chinese (zh) in our cross-lingual experiments (Table~\ref{tab:commutative-property}). Notably, the data transfer amounts exhibit non-commutative behavior, revealing variations in knowledge transfer efficiency across bidirectional language pairs.

In the English-to-Russian transfer (en, ru), data transfer is more efficient when directed from Russian to English (174.63) compared to the reverse direction (75.64), indicating an asymmetry in knowledge transfer. Similarly, in the English-to-Chinese transfer (en, zh), data transfer is more substantial from English to Chinese (66.96) than in the reverse direction (29.21).

The Russian to Chinese transfer (ru, zh) also demonstrates a non-commutative pattern, with higher data transfer from Russian to Chinese (48.47) than in the reverse direction (26.18).

The variance in mC4 subsets for each language introduces significant differences in both pretraining and evaluation datasets, potentially contributing to the absence of a commutative behavior. A more in-depth analysis would necessitate repeating experiments with equivalent datasets.

\section{Discussion}

Our study aims to measure how much knowledge, in bytes, is transferred from one language to another, enabling the investigation of the effectiveness of language-agnostic representations acquired during pretraining in cross-lingual scenarios. We hypothesize that these representations enable models to perform well on downstream tasks across diverse languages, which is observed in state-of-the-art multilingual models.

In our results, we consistently find that at least two source languages demonstrate very close $D_T$ values when evaluated against a target language, despite the diverse set of script systems and linguistic characteristics involved. This observation suggests that the data transferred from these languages to the target language is not primarily related to language-specific components but rather to language-agnostic ones. For instance, as illustrated in Figure~\ref{fig:dispersion-chart}, both English and Russian, despite being known as distant languages, achieve nearly identical $D_T$ values when evaluated on Korean, a language distinct from either. Moreover, all three source languages are remarkably close when evaluated on the Japanese test set.

Despite exposure to only a few tokens in the target language, our models demonstrate similar perplexity performance, indicating high adaptability and generalization across a broad range of languages. This reinforces the notion that the language-agnostic component plays a crucial, uniform role across source languages.

Notably, our results are not attributed to pretraining exposure to target languages, since there is a weak correlation of language contamination with the data transfer coefficient. Additionally, the observed performance is not solely dependent on language proximity, as suggested in other works.

While perplexity offers valuable insights, generalizable conclusions require evaluation in downstream tasks, especially in under-resourced languages. To explore this further, we conducted a small-scale experiment, detailed in Appendix~\ref{appendix:assin2}, finetuning our pretrained models for a low-resource language inference task. Surprisingly, we observe comparable accuracy scores between Portuguese and Russian, with good results also for English and Chinese, suggesting that our initial findings with perplexity may extend broadly. However, a more detailed investigation is required to understand cross-lingual transfer mechanisms fully.

The novelty of our approach is employing a byte-level tokenizer and adapting \citet{hernandez2021scaling} for a cross-lingual scenario. The byte-level approach facilitates consistent model embeddings across diverse scripts, enabling effective cross-lingual knowledge transfer without language-specific tokenization or preprocessing. This is supported by the strong performance of ByT5 compared to mT5 in \citet{xue-etal-2022-byt5}.

In conclusion, our study suggests the presence of language-agnostic representations contributing to cross-lingual transferability, while also laying the foundation to measure it through the Data Transfer metric. The observed consistency in model performance across diverse languages, facilitated by the byte-level tokenizer, indicates the potential for more efficient and generalizable natural language understanding across linguistic boundaries in computational linguistics and NLP.

\section{Limitations}

Our study has certain limitations that merit consideration. Firstly, our choice of initializing models with only three languages, while diverse, leaves room for improvement. Including additional languages in the pretraining phase would enhance the robustness of our analysis by minimizing possible bias towards the selected languages while providing more samples for our correlation analysis. However, this expansion would necessitate a more substantial computational budget.

Secondly, our reliance on small models, specifically a 65 million parameter model, limits the scope of our findings as larger models may exhibit different behavior. Additionally, the capacity of very large models for few-shot learning opens avenues for further exploration in the domain of transfer learning.

Lastly, the heterogeneity of the mC4 dataset across languages introduces a potential source of variability in the models' exposure to different knowledge. While the impact of this variation on data transfer remains unclear, conducting experiments with controlled datasets would offer valuable insights. Moreover, employing a more comparable test set could help mitigate statistical variance, particularly in analyses such as the commutative property assessment.

\section{Conclusion and Future Work}

Our study delves into the transferability of knowledge in cross-lingual scenarios, leveraging a byte-level tokenizer and an adapted methodology inspired by \citet{hernandez2021scaling}. By measuring the models' reliance on pretraining when executing tasks in diverse languages, our approach offers an understanding of the cross-lingual capabilities of language models. The results provide evidence that language-agnostic representations also play an important role in downstream tasks. This not only contributes to the current understanding of cross-lingual transferability but also serves as a catalyst for further exploration into the properties of language-agnostic knowledge transfer. For future research directions, we envision key investigations that can build upon the insights presented in this paper:

\begin{enumerate}
    \item \textbf{Expand Experiment Range:} Use more source languages so we can draw stronger conclusions.
    
    \item \textbf{Controlled Datasets Usage:} Employ controlled datasets and comparable test sets to address mC4 dataset heterogeneity, offering clearer insights into varied knowledge exposure impact on cross-lingual transferability and mitigating variance.

    \item \textbf{Explore Larger Models:} Investigate the use of larger models in few-shot learning downstream tasks as complementary evaluations to perplexity measurements.

    \item \textbf{Measure $D_T$ from Non-natural languages:} Perform experiments with non-natural language data, such as artificial languages with hierarchical structures. This exploration could shed light on whether knowledge transfer primarily occurs due to the content of pretraining or is largely influenced by the linguistic form.
\end{enumerate}

\section*{Acknowledgments}
We would like to thank Hugo Queiroz Abonizio, Andrea Roque, and Luciane Kawamata Notario for the useful discussions and insights. We would like to thank Google TPU Research Cloud for the computing. This research was partially funded by grant 2022/01640-2 from Fundação de Amparo à Pesquisa do Estado de São Paulo (FAPESP).

\bibliography{anthology_references,custom}

\appendix

% \section{Example Appendix}
% \label{sec:appendix}

\section{Appendix: Downstream Task Experiment}
\label{appendix:assin2}

To further explore the generalizability of our findings beyond casual language modeling with the perplexity metric, we conducted an additional experiment focusing on a distinct downstream task. This experiment involved finetuning our models pretrained on the selected source languages for a specific downstream task, targeting a different language.

\subsection{Experiment Details}

For this experiment, we selected Portuguese as the target language and the Recognizing Textual Entailment (RTE) task from the ASSIN2~\cite{real2020assin}\footnote{https://sites.google.com/view/assin2/} dataset. The dataset comprised 6,500 training examples and 500 validation instances.

We finetuned the source models for 10 epochs using a constant learning rate of $5e-5$ and a batch size of 128. The objective of the task was to predict whether one sentence entails another, with evaluation based on accuracy measured on the validation set.

\subsection{Results}

Our results, summarized in Table~\ref{tab:downstream_results}, reveal comparable performance between Portuguese (our baseline) and Russian, despite being considered a distant language. English lags by nearly 6 percentage points compared to our baseline, while Chinese exhibits the poorest performance, trailing our baseline by almost 20 percentage points.

\begin{table}[htbp]
    \centering
    \begin{tabular}{|l|c|}
        \hline
        \textbf{Source Language} & \textbf{Accuracy (\%)} \\
        \hline
        \textit{Portuguese (baseline)} & 85.0 \\
        Russian & 82.0 \\
        English & 78.6 \\
        Chinese & 65.6 \\
        \hline
    \end{tabular}
    \caption{Results from finetuning our source models on the ASSIN2 Recognizing Text Entailment task. We report the accuracy obtained in our validation set.}
    \label{tab:downstream_results}
\end{table}

Despite the limited scope of our experiment, focusing on only one target language and task, these findings suggest significant knowledge transfer across languages with varying degrees of similarity. For example, even with the lowest performance observed in Chinese, it still outperforms a random classifier by 15 percentage points (ASSIN2 contains two distinct classes).

Based on our findings, cross-lingual knowledge transfer appears to occur even with more distant languages. These results underscore the importance of further exploration in this area, indicating a promising potential for measuring knowledge transfer in cross-lingual scenarios and delineating the contributions of language-agnostic and language-specific components in the models' representations.

% This is an appendix.

\end{document}